# BIOMETRIC AUTHENTICATION USING NONPARAMETRIC METHODS

S V Sheela and K R Radhika

B M S College of Engineering, Bangalore, India

## ABSTRACT

*The physiological and behavioral trait is employed to develop biometric authentication systems. The proposed work deals with the authentication of iris and signature based on minimum variance criteria. The iris patterns are preprocessed based on area of the connected components. The segmented image used for authentication consists of the region with large variations in the gray level values. The image region is split into quadtree components. The components with minimum variance are determined from the training samples. Hu moments are applied on the components. The summation of moment values corresponding to minimum variance components are provided as input vector to k-means and fuzzy k-means classifiers. The best performance was obtained for MMU database consisting of 45 subjects. The number of subjects with zero False Rejection Rate [FRR] was 44 and number of subjects with zero False Acceptance Rate [FAR] was 45. This paper addresses the computational load reduction in off-line signature verification based on minimal features using k-means, fuzzy k-means, k-nn, fuzzy k-nn and novel average-max approaches. FRR of 8.13% and FAR of 10% was achieved using k-nn classifier. The signature is a biometric, where variations in a genuine case, is a natural expectation. In the genuine signature, certain parts of signature vary from one instance to another. The system aims to provide simple, fast and robust system using less number of features when compared to state of art works.*

## 1. INTRODUCTION

Biometric authentication technique based on the pattern of human iris is well suited to be applied to any access control system that requires high level of security. Iris is the annular ring between the pupil and the sclera of the eye. The unique texture pattern is used as biometric signature. The structural formation in the human iris remains constant over time, there by exhibits long-term stability. The variations in the gray level intensity values effectively distinguish two individuals. The iris recognition techniques provide authorized access to ATMs, credit cards, desktop PCs, workstations, buildings and computer networks. Iris based security systems capture iris patterns of individuals and match these patterns against record in available databases. Even though significant progress has been made in iris recognition, handling noisy and degraded iris images require further investigation. The iris recognition algorithms need to be developed and tested in diverse environment configurations. Most of the existing methods have limited capabilities to recognize features in realistic situations. The challenging concepts are based on iris localization, nonlinear normalization, occlusion, segmentation, liveness detection and large scale identification. In off-line signature verification system, the signatures are treated as gray level images. The image can be a saved file acquired by a tablet or can be scanned from a copy of document [1]. The features are to be invariant to rotation, translation and scaling of the object sample [2]. Some of the static features are vertical midpoints, number of vertical midpoint crossings in signature, total pen travel writing distance, signature area, maximum pixel change. In this paper, subpatterns of signature area are considered under minimum variance criteria. It is well known that no two genuine signatures of a person are precisely the same and some signature experts note that if two signatures written on paper were same, then they could be considered as forgery by tracing [3]. The technique eliminates the subpatterns with more variance in a signature after learning from genuine training samples.

The *k*-means clustering is a method of cluster analysis which aims to partition *n* observations into *k* clusters in which each observation belongs to the cluster with the nearest mean. Given a set of observations $(x_1, x_2, ..., x_n)$, where each observation is a d-dimensional real vector, *k*-means clustering aims to partition the *n* observations into *k* sets $(k < n)$ $S=\{S_1, S_2,..., S_k\}$ so as to minimize the within-cluster sum of squares given by Equation (1).

$$\arg\min_{S} \sum_{i=1}^{k} \sum_{x_j \in S_i} \|x_j - \mu_i\|^2 \qquad (1)$$

where $\mu_i$ is the mean of $S_i$.

The segmented iris images are termed as Pupil Iris Frame [PIF] images. The k-means clustering is used in classification of the PIF images. Given a set of PIF images, the clustering algorithm classifies each image as belonging to a particular cluster. The algorithm returns the *k* cluster indices in vector *c* and centroid locations in the matrix *f*. In the experiment, the genuine and imposter PIF images are considered for training. The PIF images are classified into 2 clusters, genuine and imposter. An instance of *c* and *f* values are given as *c* = [ 1 1 1 1 1 1 1 2 2 2] and *f* = [0.9236 0.0486]. The *c* array indicates that the first seven images belong to the first cluster and remaining three belong to the second. The sequence of steps developed for implementing the *k*-means clustering is given in Algorithm 1.

Algorithm 1: *k*-means clustering
Let n be the number of PIF images and k be the number of clusters.
1. The values of *n* and *k* are initialized.
2. Moment summation value is computed for training samples.
3. The *k*-means clustering algorithm is implemented. The initial centroids are computed.
4. The algorithm assigns the sample to the cluster with closest centroid. The centroids are recalculated.
5. Step 4 is repeated until there is no change in centroid values. The algorithm returns *k* cluster indices and centroid locations.

In fuzzy clustering each sample has some graded or fuzzy membership in a cluster. The fuzzy *k*-means clustering algorithm attempts to partition a finite collection of elements $X=\{x_1,x_2,....,x_n\}$ into a collection of *k* fuzzy clusters with respect to some given criterion. Given a finite set of data, the algorithm returns a list of *k* cluster centres V, such that $V = v_i$, $i =1, 2, ... , k$ and a partition matrix U such that $U = u_{ij}$, $i =1, ..., k$, $j =1,..., n$ where $u_{ij}$ is a numerical value in [0, 1] that tells the degree to which the element $x_j$ belongs to the $i^{th}$ cluster. This method allows the data to belong to two or more clusters. It is based on minimization of the objective function. The objective function is given by Equation (2).

$$J_m = \sum_{i=1}^{n} \sum_{j=1}^{c} u_{ij}^{m} \|x_i - v_j\|^2, 1 \leq m \leq \infty \qquad (2)$$

Fuzzy partitioning is carried out through an iterative optimization of the objective function with the update of membership and cluster centres. The termination condition is given by Equation (3).

$$\max_{ij} \left\{ \left| u_{ij}^{(k+1)} - u_{ij}^{(k)} \right| \right\} < \varepsilon \qquad (3)$$

where $\varepsilon$ is the termination criterion and $k$ is the number of iteration steps.

The moments are computed for the PIF images. The fuzzy clustering algorithm is applied for the moments. In the proposed system, the number of clusters $c$ is 2. The algorithm returns the partition matrix or membership function matrix $U$ which indicates the degree of membership. An instance of partition matrix is shown in Table 1.

Table 1. Partition matrix

| Image | Membership value for Cluster 1 | Membership value for Cluster 2 | Classification cluster (Inference) |
|---|---|---|---|
| 1 | 0.037851 | 0.962149 | 2 |
| 2 | 3.61E-05 | 0.999964 | 2 |
| 3 | 0.027781 | 0.972219 | 2 |
| 4 | 0.00341 | 0.99659 | 2 |
| 5 | 0.01285 | 0.98715 | 2 |
| 6 | 0.012207 | 0.987793 | 2 |
| 7 | 0.001298 | 0.998702 | 2 |
| 8 | 0.999855 | 0.000145 | 1 |
| 9 | 1 | 1.11E-07 | 1 |
| 10 | 0.999903 | 9.72E-05 | 1 |

The maximum value in the partition matrix for each cluster is determined. This indicates the degree of membership to a particular cluster. For example, in Table 1, the degree of membership of the first PIF image is 0.037851 for cluster 1 and 0.962149 for cluster 2. The maximum membership value indicates that the image belongs to cluster 2. The index of these values is used to count the number of images belonging to each cluster. The sequence of steps for implementing Fuzzy $k$-means clustering is given in Algorithm 2.

Algorithm 2: Fuzzy $k$-means clustering

Let $n$ be the number of training samples and $k$ be the number of clusters.
1. The values of $n$ and $k$ are initialized.
2. Moment summation value is computed for training samples.
3. The fuzzy clustering algorithm is implemented. The initial centroids are computed using moment summation values of $P$ training samples. The iterative update of centroid values take place for every insertion to the cluster. The insertion to a cluster is based on Euclidean distance measure. The algorithm returns the partition matrix $U$.
4. The maximum value in the partition matrix is determined.
5. The index of the maximum value is used in classification.

In classification problems, complete statistical knowledge regarding the conditional density functions of each class is rarely available, which precludes application of the optimal Bayes classification procedure. When no evidence supports one form of the density functions rather than another, a good solution is often to build up a collection of correctly classified samples, called the training set, and to classify each new pattern using the evidence of nearby sample observation. One such non-parametric procedure has been introduced by Fix and Hodges and has since become well-known in the pattern recognition literature as the voting $k$-nearest neighbour ($k$-nn) rule [4]. Conceptually, a $k$-nn classification algorithm has two independent sections. They are, minimal consistent subset selection section and finding the $k$-nearest neighbor for an unseen object. The tolerant rough set or evidence theory can be used to select a set of objects from the training data that have the same classification power as the original data set. This eliminates irrelevant and redundant attributes providing insight into the relative significance of the samples in the training set [5,7,8]. The shortcomings of *k-nn* are each neighbor is equally important, prone to be affected by the imbalanced data problem and necessity of keeping the whole reference set in the computer memory. Large classes always have a better chance to win [9]. In modern computer era, fast learning and error rate estimation by leave one out method, makes *k-nn* significantly useful [10]. For the *k-nn* approach, *k* is the square root of the number of learning instances [6]. By the fuzzy *k-nn* algorithm, the criteria for the assignment of membership degree to a new object depend on the closeness of the new object to its nearest neighbors and the strength of membership of these neighbors in the corresponding classes. The advantages lie both in the avoidance of an arbitrary assignment and in the support of a degree of relevance from the resulting classification.

The material is organized in following manner. In the next section, preliminaries of quadtree, Hu moments, variance and the classifiers are explained. Following that, the state of art is discussed. In fourth section proposed system is presented. The fifth section consists of experimental results and comparison of the approaches. The last section concludes with a discussion of findings.

## 2. PROLOGUE

A tree data structure in which each internal node has up to four children is termed as quad tree. The input space is decomposed into adaptable cells. The tree directory follows the spatial decomposition of the quadtree. Quadtrees are classified according to the type of data they represent, including areas, points, lines and curves. In this work region quadtrees are used. The region quad tree represents a partition of space in two dimensions by decomposing the region into four equal quadrants. A region quadtree with a depth of *n* may be used to represent an image consisting of $2n \times 2n$ pixels, where each pixel value is 0 or 1. The root node represents the entire binary image. Let *R* represent the entire normalized binary signature image. Quadtree partitions R into 4 sub regions, $R_1, R_2, R_3, R_4$, such that (a) $\cup R_i = R$. (b) $R_i$ is a connected region, $i=1,2,3,4$ (c) $R_i \cap R_j = \Phi$ for all *i* and *j*, $i \neq j$. For further subdivisions same clauses are applicable. The proposed system implements first, second and third trie level of decomposition to represent variable size of the normalised binary signature. This data structure is selected because, in a real time scenario storing in external storage files are simpler since every node is either a leaf or it contains exactly four children as compared to binary trees which involve number of traversals with level numbers of nodes for different encode levels. In the image processing field, centroid and size normalization provide significant inferences even in on-line scenarios [11]. Orientation independence is achieved by orthogonal moment invariant to a pair of uniquely determined principal axes to characterize each pattern for recognition [12]. Moments of order *p, q* of a binary image *I* are calculated as given in Equation (4).

$$m_{pq} = \sum_{i,j \in I} i^p j^q \qquad (4)$$

Hu derived moment expressions is extended as centralized moments given by Equation (5).

$$M_{pq} = \sum_{i,j \in I} (i-a)^p (j-b)^q \qquad (5)$$

These are invariant to translation, rotation and scaling of shapes. The parameters *a* and *b* are the centers of mass in the 2D co-ordinate system. Centralized moments are invariant to translations of the image. This is equivalent to moments of an image that has been shifted such that the image centroid coincides with the origin [23]. The lower order moments derive the shape characteristics. The first, second and eighth Hu moments are used in this work. The moments from 3 to 7, are usually assigned to moment invariants of order 3 are not considered [11]. For the entire material, these moments are identified as $Moment_A$, $Moment_B$ and $Moment_C$. The expressions of three moments are given by Equations (6)-(8) respectively.

$$Moment_A = \frac{M_{20} + M_{02}}{(m_{00})^2} \tag{6}$$

$$Moment_B = \frac{(M_{20} - M_{02})^2 + 4M_{11}^2}{(m_{00})^2} \tag{7}$$

$$Moment_C = \frac{M_{20}M_{02} - M_{11}^2}{(m_{00})^4} \tag{8}$$

Authentication via moment based descriptors is achieved through variance criteria. Let $M$ be the order of the normalised binary image $I$. In this work $M = 512$ is considered. $L$ denotes number of subregions formed on the application of quadtree procedure on $I$. The value of $L$ is computed using Equation (9).

$$L = \frac{M}{d_1} X \frac{M}{d_1} \tag{9}$$

where $d_1 = 64, 128, 256$. $d_1$ denotes the minimum subregion size which forms $M/d_1$ trie level of quadtree. Moments are applied on each of the $L$ quadtree components [QCs]. The variance of each of corresponding subregion in $P$ genuine samples from the training set is found as shown in Figure 1. The average variance is calculated. Each $var_i$, $i \in \{1,..,L\}$ less than average is selected for subregion list. Let $b$ be the threshold parameter of the system. If the number of elements in the subregion list is greater than $b$, process is repeated with new average of variance for the subregions in the list. The $b$ template MVQCs are obtained which denote less variation subregions of signature of a person with respect to moment applied.

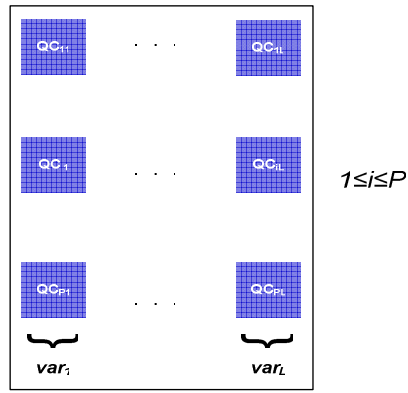

Figure 1. Depict subregion-wise variance calculation for trie level of $M/d_1$.

## 3. STATE OF ART

The first iris recognition system developed by J.Daugman was based on phase-based approach [24]. The representation of iris texture is binary coded by quantizing the phase response of a texture filter using quadrature 2D Gabor wavelets into four levels. Iris codes are generated and Hamming Distance is used as a measure of dissimilarity. The Equal Error Rate [EER] of 0.08 was obtained. Continuing the Daugman's method, Karen Hollingsworth, Kevin Bowyer and Patrick Flynn [25] has developed a number of techniques for improving recognition rates. These techniques include fragile bit masking, signal-level fusion of iris images, detecting local distortions in iris texture and analysing the effects of pupil dilation. The experiments were conducted on ICE database. The Hamming Distance of 7.48 and 0.15 was obtained for fragile bit masking and local distortion detection techniques. The EER of 0.0038 and 0.068 was obtained for signal-level fusion and analysis of pupil dilation effects. The system developed by Wildes is based on texture analysis [26]. The Laplacian of Gaussian (LoG) is applied to the image at multiple scales and the resulting Laplacian pyramid constructed with different levels serves as basis for further processing. EER of 1.76 was obtained in this method. Iris recognition system developed by Li Ma is characterized by local intensity variations [27]. The sharp variation points of iris patterns are recorded as features. The feature extraction generates 1D intensity signals considering the information density in the angular direction. The feature values are the mean and the average absolute deviation of the magnitude of each 8x8 block in the filtered image. The correct recognition rate of 94.33% was obtained. The method by Li Ma was further improved by Zhenan Sun [28] where in the local feature based classifier was combined with an iris blob matcher. The blob matching aimed at finding the spatial correspondences between the blocks in the input image and that in the stored model. The similarity is based on the number of matched block pairs. The block attributes are recorded as centroid coordinates, area and second order central moments. H. Proenca and L.A.Alexandre [29] proposed a moment-based texture segmentation algorithm, using second order geometric moments of the image as texture features. The clustering algorithms like self-organizing maps, *k*-means and fuzzy *k*-means were used to segment the image and produce as output the clusters-labelled images. The experiments were conducted on UBIRIS database with accuracy of 98.02% and 97.88% for images captured in session 1 and session 2, respectively. The work proposed by Nicolaien Popescu for iris segmentation uses k-means quantization to determine crisp and fuzzy iris boundaries. The experiment was conducted on Bath University iris database with 6 segmentation failures [33]. Iris recognition system by Wen-Shiung Chen is based on wavelet transformation [34]. K-means classifier was used for recognition. EER of 11.3%, 9.7%, 10.5% and 9.3% were obtained considering four different measures such as contrast, correlation, homogeneity and entropy respectively.

Many of the verification systems use writer dependent threshold and writer independent thresholds. The recognition system using warping proposed by Gady Agam [1] is with the dataset built by scanned documents. Signatures of 76 subjects with each of 5 samples in test collection were extracted. The approach obtained rates of 100% precision with 30% recall. In the verification system using enhanced modified direction features proposed by Vu Nguyen [14], the classifiers were trained using 3840 genuine and 4800 targeted forged samples. FARR is the measurement of false acceptance rate for random forgery and FARG is measurement of false acceptance rate for targeted forgery. DER is distinguishing error rate, which is average of FARR and FARG. The system obtained DER of 17.78% with SVM. FARR for random forgeries was below 0.16%. The system for fuzzy vault construction proposed by Manuel [15] used MCYT, Spain database for training. The system achieved seperability distance of 12 for random forgeries. The distance is termed as average distance between genuine and impostor vault input vectors. The work proposed by Edson [16] for a real application (4-6 samples), the results presented for false rejection error rate was 13% for HMM. In Hanmandlu's approach [17] using TS model with consequent coefficients fixed with his second formulation (which depends on number of rules) out of 200 genuine signatures 125 were accepted.

## 4. PROPOSED SYSTEM

### 4.1 Preprocessing

The preprocessing of iris images is based on the area of connected components. The connectivity between pixels in a gray scale image is determined based on certain conditions of gray level values and spatial adjacency [22]. For a pixel e with the coordinates *(x,y)* the set of pixels given by *D(e) ={(x+1,y), (x-1,y), (x,y+1), (x,y-1), (x+1,y+1), (x+1,y-1), (x-1,y+1), (x-1,y-1)}* is called its 8-neighbors. A connected component is a set of pixels in an image which are all connected to each other. All pixels in a connected component share same set of intensity values. The process of extracting various disjoint and connected components in an image and marking each of them with a distinct label is called connected component labelling.

The histogram of the eye image is the plot of number of pixels corresponding to each gray level value in the range [0,255]. The highest peak in the pupil area corresponds to count of pixels with gray level value nearer to zero. The gray level value corresponding to this peak is given by *ind*. Consider all pixels less than or equal to *ind*. This results in a binary image *bw*. The steps are represented using Equations (10)-(13).

$$(count, bins) = histogram(I(x,y)) \quad (10)$$
$$maxcount = max(count) \quad (11)$$
$$ind = bins(maxcount) \quad (12)$$
$$bw = I(x,y) <= ind \quad (13)$$

The array count contains the number of pixels for each gray scale value in bins. All the connected components in *bw* are labelled using Equation (14). The labelling is based on 8-connected components. The label 0 indicates background. The label 1 corresponds to component 1, label 2 corresponds to component 2 and so on. The total number of connected components is given by *num*.
$$(F, num) = label(bw) \quad (14)$$
*F* is the matrix containing the labels. The area of all the connected components is determined using Equation (15).
$$Comp(i) = area(F==i) \quad i=1,2,...num \quad (15)$$

The array *Comp* consists of the area of connected components. The estimation of area is based on the number of nonzero pixels. There exist several components with smaller area and one component with larger area. The component with larger area corresponds to the pupil. This is achieved by finding the maximum of all the areas using Equation (16).
$$maxarea = max(Comp) \quad (16)$$
*maxarea* corresponds to the circular or elliptical area with pixels that determines the shape of the pupil. The radius of the area is determined along both horizontal and vertical directions using Equations (17) and (18).

$$radius_1 = (x_{max} - x_{min})/2 \quad (17)$$
$$radius_2 = (y_{max} - y_{min})/2 \quad (18)$$
where $x_{max}$, $x_{min}$ are the maximum and minimum *x* coordinate values. $y_{max}$ and $y_{min}$ are the maximum and minimum *y* coordinate values. The coordinates correspond to the largest connected component which is circular or elliptical. The center coordinates of the pupil is determined using Equations (19) and (20)

$$x_c=(x_{max}+x_{min})/2 \quad (19)$$
$$y_c=(y_{max}+y_{min})/2 \quad (20)$$

Using the center coordinates an appropriate window size is determined. The window size is the basis to clearly distinguish two subjects based on moment summation values. A range of window sizes was worked out and it was found that for a particular window and number of quadtree components, the moment summation values is similar for genuine samples and different for imposter samples. Storage of eye images requires more memory space for large databases. The segmented iris images for particular window size would drastically reduce the storage capacity. The segmented iris images with an accurate window size are the PIF images. The size of each PIF image is in the range of 1.5KB to 4KB as compared to eye images of size 300KB to 400KB.

**Window Selection**

With $(x_c, y_c)$ as center coordinates a bounded rectangle or window is determined. Initially, the length and breadth of the rectangle was equal to radius where $radius=max(radius_1, radius_2)$. The moment summation values were indistinguishable for values of $b$ considered. $b$=16, 10, 8 and 6. The offset values were added both length-wise and breadth-wise incrementally. The bounded rectangle is a four-element position vector defined by $x$-value, $y$-value, width and height to determine the size and position.

$$x\text{-value}=y_c\text{-radius-offset}_1 \quad (21)$$
$$y\text{-value}=x_c\text{-radius-offset}_1 \quad (22)$$
$$width=2*radius+offset_2 \quad (23)$$
$$height=2*radius+offset_2 \quad (24)$$

The preprocessing stages are shown in Figure 2 and the quadtree numbering is shown in Figure 3.

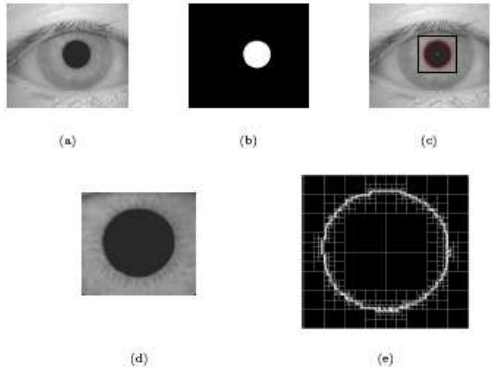
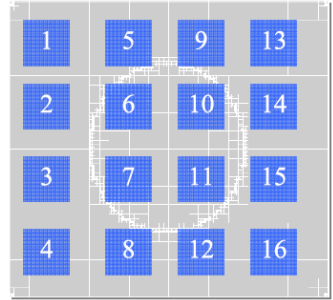

Fig. 2 Preprocessing Stages      Fig.3 Quadtree tiles

In the proposed work, the size of the PIF image is resized to 512 x 512. $d_1$=128, 256 are considered. Moments are applied for each QC. The value of $b$ determines the number of QCs which clearly distinguishes the moment summation values between two subjects. The variance of each of the corresponding QCs in $P$ training samples is determined. The average variance is calculated. Each QC with variance less than average variance is listed. The computation of average variance and determining the number of QCs less than average is repeated for a particular value of $b$. The list of $b$ QCs denotes less variation iris subregions. An instance of QCs with minimum variance for a subject is given in Tables 2-5 for $b$=16, 10, 8 and 6 respectively.

Table 2. Minimum Variance Calculation for $b$=16

### Table 3. Minimum Variance Calculation for $b=10$

| Sample | 1 | 2 | 3 | 4 | 5 | 6 | 7 | 8 | 9 | 10 | 11 | 12 | 13 | 14 | 15 | 16 | avg |
|---|---|---|---|---|---|---|---|---|---|---|---|---|---|---|---|---|---|
| 1 | 1.66 E+39 | 6.80 E+38 | 1.46 E+39 | 8.50 E+39 | 3.05 E+38 | 1.27 E+19 | 2.62 E+22 | 4.60 E+39 | 1.58 E+38 | 5.00 E+16 | 2.56 E+30 | 1.64 E+40 | 2.60 E+39 | 4.43 E+39 | 1.82 E+40 | 2.11 E+40 | |
| 2 | 2.28 E+39 | 9.07 E+38 | 2.41 E+39 | 1.44 E+40 | 1.82 E+38 | 1.20 E+18 | 2.82 E+19 | 3.98 E+39 | 1.08 E+38 | 3.68 E+18 | 1.38 E+26 | 9.86 E+39 | 7.81 E+38 | 8.24 E+38 | 9.71 E+39 | 1.74 E+40 | |
| 3 | 3.66 E+39 | 1.34 E+39 | 5.45 E+39 | 2.03 E+40 | 1.35 E+38 | 2.32 E+16 | 1.69 E+24 | 6.91 E+39 | 4.94 E+37 | 5.17 E+18 | 1.32 E+29 | 1.57 E+40 | 7.83 E+38 | 6.62 E+38 | 1.02 E+40 | 1.42 E+40 | |
| min-var | 1.048 E+78 | 1.116 E+77 | 4.353 E+78 | 3.492 E+79 | 7.684 E+75 | 4.914 E+37 | 9.374 E+47 | 2.383 E+78 | 2.927 E+75 | 6.924 E+36 | 2.084 E+60 | 1.299 E+79 | 1.098 E+78 | 4.530 E+78 | 2.276 E+79 | 1.219 E+79 | 6.025 E+78 |

### Table 4. Minimum Variance Calculation for $b=10$    Table 5. Minimum Variance Calculation for $b=10$

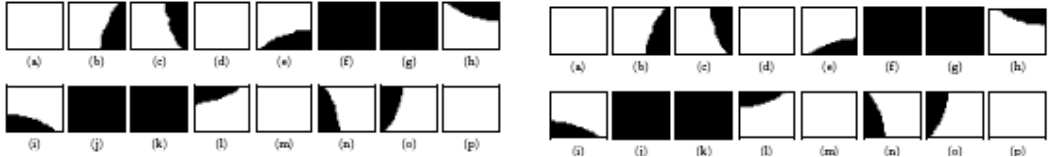

| Sample | 1 | 2 | 5 | 6 | 7 | 9 | 10 | 11 | 13 | avg |
|---|---|---|---|---|---|---|---|---|---|---|
| 1 | 1.66 E+39 | 6.80 E+38 | 3.05 E+38 | 1.27 E+19 | 2.62 E+22 | 1.58 E+38 | 5.00 E+16 | 2.56 E+30 | 2.60 E+39 | |
| 2 | 2.28 E+39 | 9.07 E+38 | 1.82 E+38 | 1.20 E+18 | 2.82 E+19 | 1.08 E+38 | 3.68 E+18 | 1.38 E+26 | 7.81 E+38 | |
| 3 | 3.66 E+39 | 1.34 E+39 | 1.35 E+38 | 2.32 E+16 | 1.69 E+24 | 4.94 E+37 | 5.17 E+18 | 1.32 E+29 | 7.83 E+38 | |
| min-var | 1.048 E+78 | 1.116 E+77 | 7.684 E+75 | 4.914 E+37 | 9.374 E+47 | 2.927 E+75 | 6.924 E+36 | 2.084 E+60 | 1.09 E+78 | 2.521 E+77 |

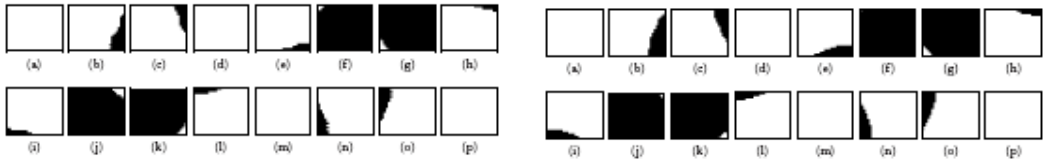

| Sample | 2 | 5 | 6 | 7 | 9 | 10 | 11 | avg |
|---|---|---|---|---|---|---|---|---|
| 1 | 6.80 E+38 | 3.05 E+38 | 1.27 E+19 | 2.62 E+ | 1.58 E+38 | 5.00 E+16 | 2.56 E+30 | |
| 2 | 9.07 E+38 | 1.82 E+38 | 1.20 E+18 | 2.82 E+19 | 1.08 E+38 | 3.68 E+18 | 1.38 E+26 | |
| 3 | 1.34 E+39 | 1.35 E+38 | 2.32 E+16 | 1.69 E+24 | 4.94 E+37 | 5.17 E+18 | 1.32 E+29 | |
| min-avg | 1.116 E+77 | 7.684 E+75 | 4.914 E+37 | 9.374 E+47 | 2.927 E+75 | 6.924 E+36 | 2.084 E+60 | 1.746 E+76 |

| Sample | 5 | 6 | 7 | 9 | 10 | 11 | avg |
|---|---|---|---|---|---|---|---|
| 1 | 3.05 E+38 | 1.27 E+19 | 2.62 E+22 | 1.58 E+38 | 5.00 E+16 | 2.56 E+30 | |
| 2 | 1.82 E+38 | 1.20 E+18 | 2.82 E+19 | 1.08 E+38 | 3.68 E+18 | 1.38 E+26 | |
| 3 | 1.35 E+38 | 2.32 E+16 | 1.69 E+24 | 4.94 E+37 | 5.17 E+18 | 1.32 E+29 | |
| min-avg | 7.684 E+75 | 4.914 E+37 | 9.374 E+47 | 2.927 E+75 | 6.924 E+36 | 2.084 E+60 | 1.768 E+75 |

The QCs of genuine and imposter samples are shown in Figures 4 and 5. The MVQCs for $b=6$ are shown in Figure 6. The first two rows are MVQCs of genuine samples. For instance the MVQCs are {5, 6, 7, 9, 10, 11}. The next two rows are the QCs in imposter corresponding to these MVQCs.

Figure 4. Quadtree Components of two genuine samples

Figure 5. Quadtree Components of two imposter samples

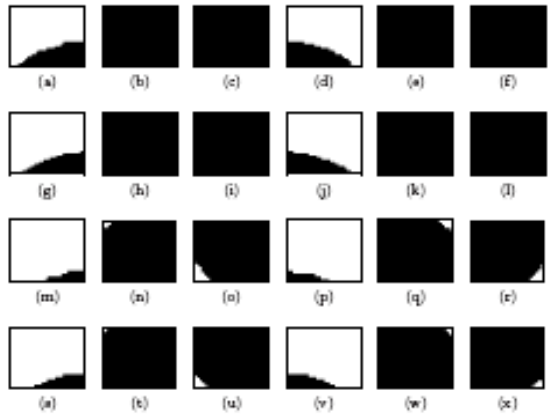

Figure 6. Minimum Variance Quadtree Components {5,6,7,9,10,11}

The summation values of moments for MVQCs are computed. It is observed that the moment summation values are different for genuine and imposter iris samples. This is shown in Figure 7.

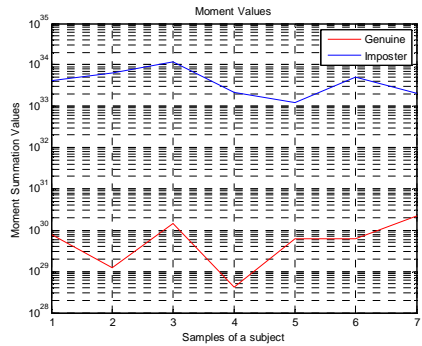

Figure 7. Moment Summation Values for iris samples

The signature image of size 850x360 is read and is converted to gray from rgb format. To maintain the uniformity the sample is resized for 512x512. The signature is normalized to minimum bounding box. Quadtree is constructed for 2, 4, 8 depth levels.

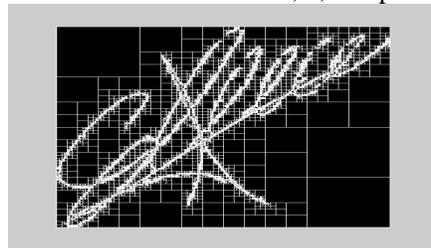

Figure 8. Quadtree decomposition of a sample

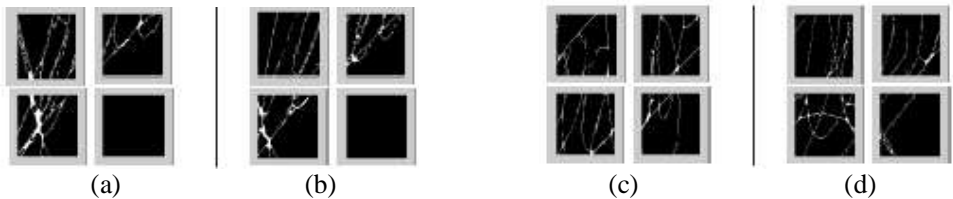

Figure 9. (a)-(b) Four blocks of genuine samples of a subject (c)-(d) Four blocks of imposter samples of a subject

The Figure 8 depicts the 4 trie level and instances of quadtree components are shown in Figure 9. The geometric moments are applied for quadtree components. The procedure is explained for the case of trie level = 4, which depicts minimum size of the component to be 128 x 128. The $L$ = 16 components formed are numbered. The $P$ training samples of the person are considered. The sixteen components are subjected to the moments. The variance of corresponding quadtree components for $P$ training samples are calculated. $b$ is the number of MVQC's are selected. Figures 10 and 11 show the quadtree decomposition, minimum variance and maximum variance quad tree components of genuine samples for $b = 2$. The first two rows in Figure 11 depicts arrangement of quadtree components of two genuine samples, where as last two rows depict imposter samples. The minimum variance components observed in genuine samples are noted. The corresponding components in imposter samples are not minimum variance components. The $b$ standard template quadtree components are found for a subject. This signifies that hand written signature varies from one instance to another. It is a natural variation. In this work, standard template quad tree components found for a subject are used to detect imposter signatures. The MVQC's variation is not minimum for an imposter sample.

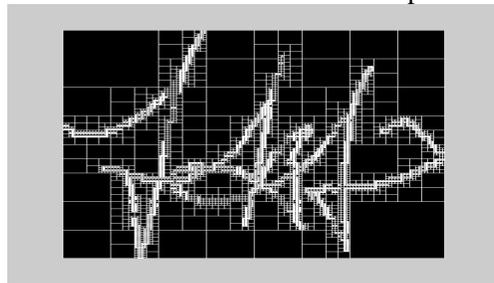

Figure 10. Genuine sample of a subject

For one of the subject in the database, for $b = 4$, the MVQCs listed were 2, 3, 13, 15 with $Moment_C$ applied. This is a learning obtained by using only genuine $P$ training samples of the person. The summation of moment value on all MVQCs is calculated. In many cases, the summation values in imposter sample were larger when compared with summation in genuine sample. Experimentally it was evident that, MVQC list for a particular $b$ value is different for different moment types. The $P = 10$ genuine and imposter training samples for respective MVQC summation values are plotted for a subject with $b = 4$ and $d_1 = 128$. This is shown in Figure 12 (a) - (f).

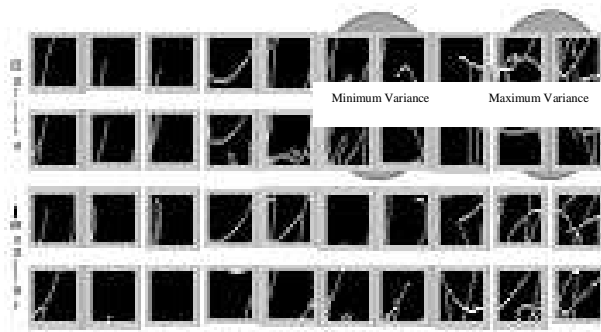

Figure 11. Depicts minimum variance and maximum variance parts in genuine samples

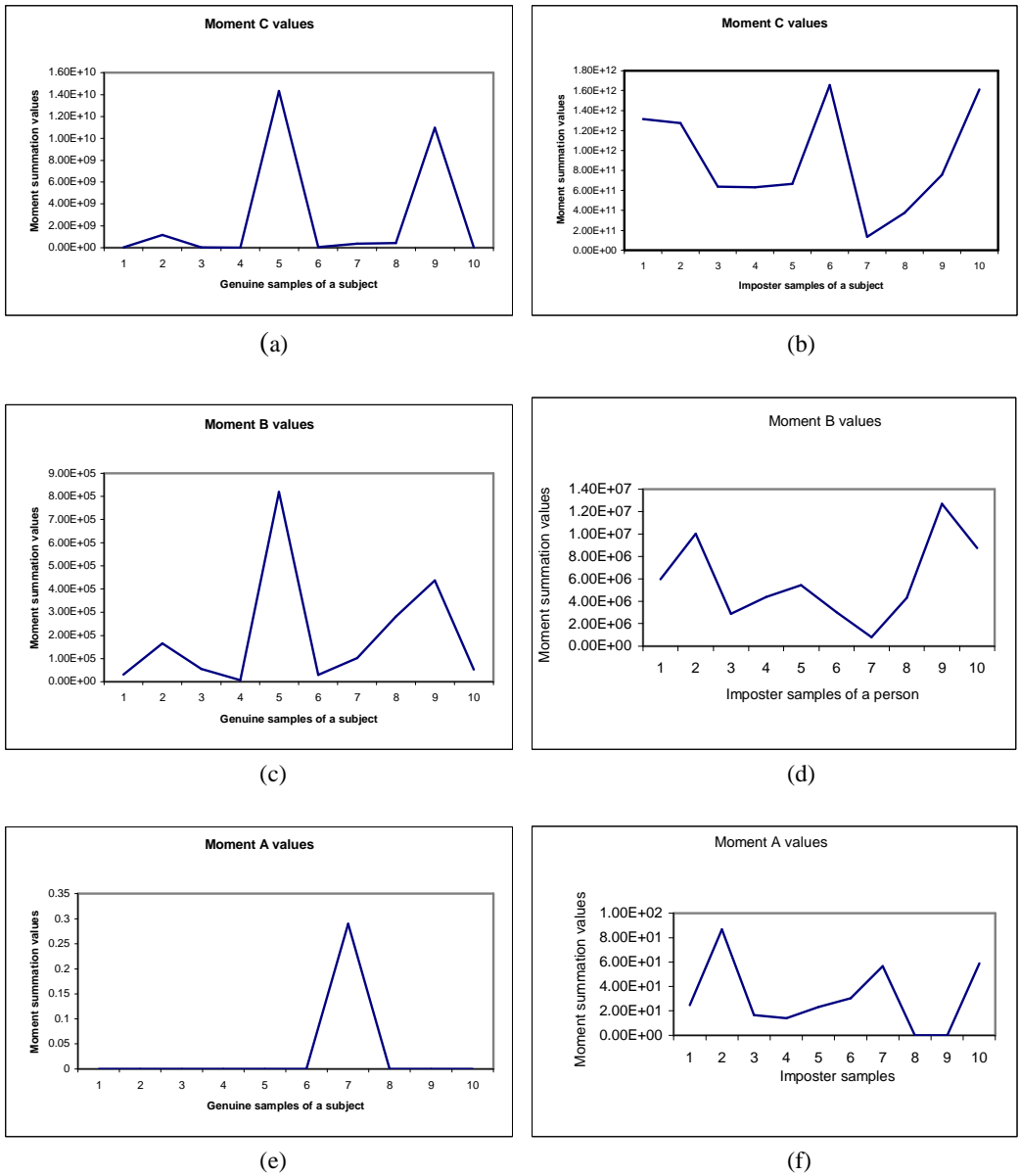

Figure 12. Moment Summation Values for signature samples

## 4.2. k-means Method

The moment values are higher for imposter samples as compared to genuine samples. Suitable to the above observation, k-means as heuristic method produce optimal result as shown in Figure 13. Fuzzy k-means which is more suitable for subpattern analysis also produce satisfactory results.

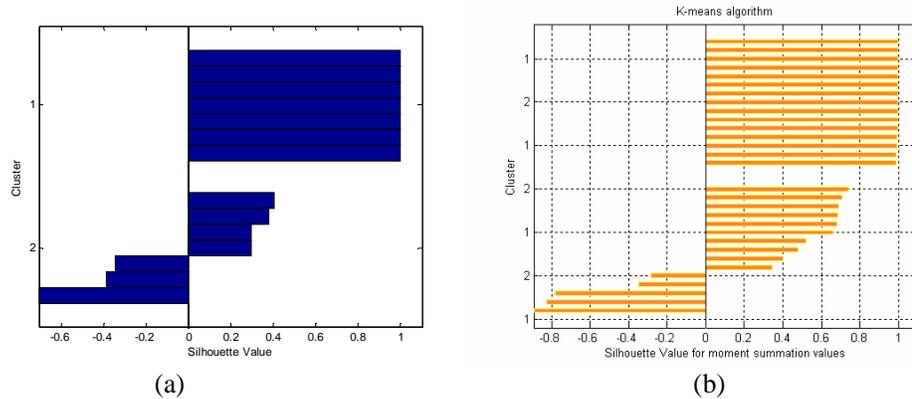

(a)                          (b)

Figure 13. Silhouette diagram implemented with city block distance for (a) iris (b) signature

### 4.3. Criteria for initial mean selection

Let $H$ be the set which represents the moment summation values of $P$ training samples. $H=\{s_1,s_2,\ldots s_P\}$. If $m_1$ is minimum value in $H$ and $m_2$ is maximum value in $H$, subsequently threshold for selecting the second initial centroid will be a value greater than $m_2$. Initial centroids are selected using Equations (25) and (26). Figure 14 depicts the initial centroid selection.

$$initial\_centroid_1 = m_1$$
$$initial\_centroid_2 = (m_1 + threshold)/2$$
(25)-(26)

The $P$ genuine training samples considered provide the criteria for initial centroids selection based on moment summation values. The experiment takes into account the minimum variance blocks observed using m genuine training samples. There is unambiguous separation between moment summation values of genuine and imposter samples.

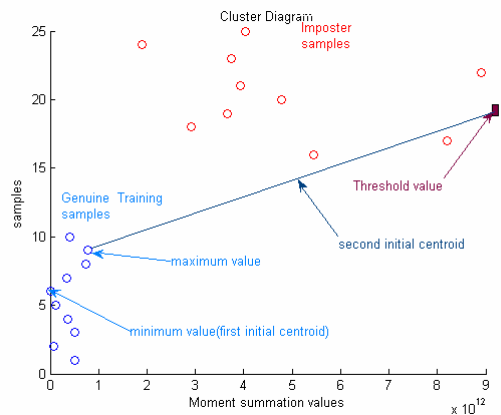

Figure 14. Initial Centroid Selection

### 4.3. *k-nn* method

Algorithms for speeding-up the *k-nn* search fall into two categories: template condensation and template reorganization. Template condensation removes redundant patterns in a template set and template reorganization, restructures templates for efficient search of *k* nearest neighbors [19]. In this work, template condensation is achieved by genuine training samples selected. Template reorganization is achieved with the aid of local mean, in training session. In fuzzy k-nn, the initial membership value equal to 1 is initialized to genuine training samples. Based on values obtained for membership values, testing samples are classified.

### 4.4. Averagemax Method

Since the feature extraction was very much successful, the mean of the moment summation values of genuine training samples will provide an optimistic threshold for classification. Furthering observing, the maximum difference value of each term of *H* with average value, provided a threshold for classification. The threshold factor is given by Equation (27).

$$factor = \max\left\{H(i) - \frac{\sum_{j=1}^{P} H(j)}{P}\right\} i \in \{1, 2, ..., P\} \quad (27)$$

### 5. Experiments

The experiments were conducted on three iris databases, Chinese Academy of Sciences Institute of Automation (CASIA) [30], Iris Challenge Evaluation (ICE) [31] and Multimedia University (MMU) [32] databases. The CASIAv1 database consists of 756 eye images collected from 108 subjects with 7 samples per subject. A subset of ICE database consisting of 623 eye images from 89 subjects was used considering 7 images per subject. The left eye images from MMU database was used with a total of 225 images from 45 subjects with 5 images per subject. The number of training samples P=3 for all the three databases. For the three databases, the window size in terms of (offset$_1$, offset$_2$) is determined which gives the best distinguishing moment summation values for different subjects. At *b*=16, it is observed that the moment summation values cannot be distinguished for most of the samples.

Table 6. Window Selection at $d_1$=128

| Database | (offset$_1$, offset$_2$) | b |
|---|---|---|
| CASIA | (20,40) | 10 |
| ICE | (6,12) | 6 |
| MMU | (20,40) | 8 |

At $d_1$=256, the experiment was conducted for *b* = 2, 3. At $d_1$=128, the experiment was conducted for *b*= 6, 8, 10. The plot of average FRR and FAR values are shown in Figures 15, 16 and 17.

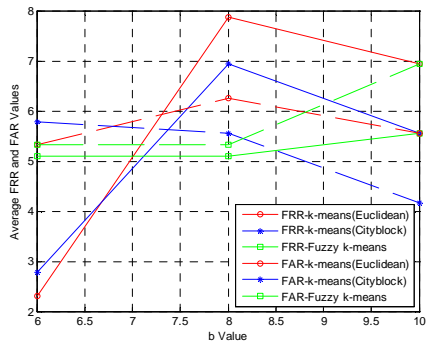
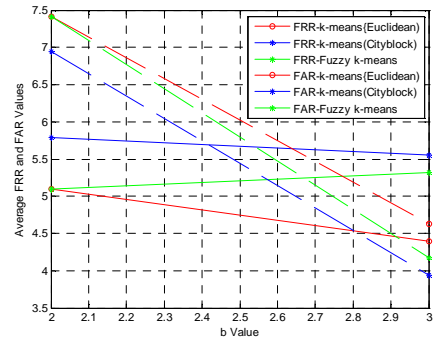

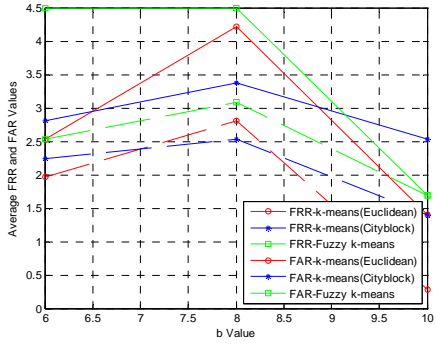 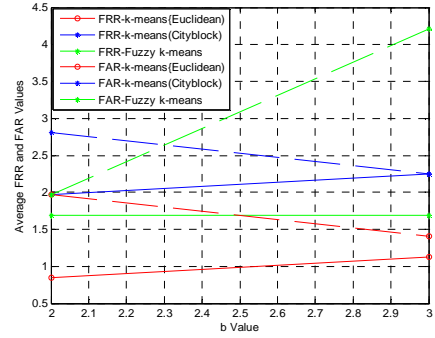

(a)                  (b)

Figure 15. Average FRR and FAR for CASIA database with (a) $d_1$=128 (b) $d_2$=256

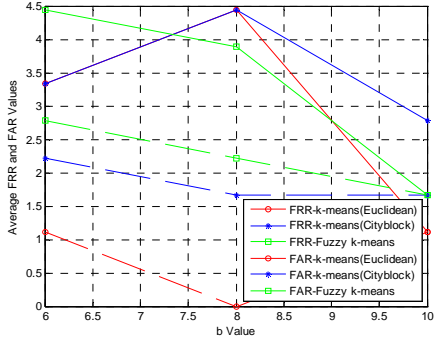 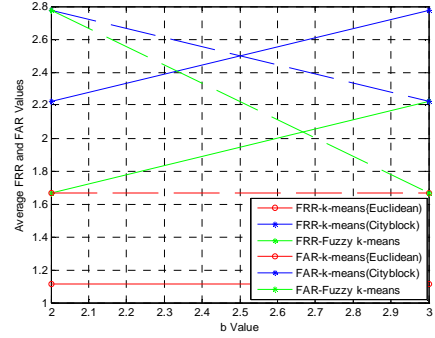

(a)                  (b)

Figure 16. Average FRR and FAR for ICE database with (a) $d_1$=128 (b) $d_2$=256

(a)                  (b)

Figure 17. Average FRR and FAR for MMU database with (a) $d_1$=128 (b) $d_2$=256

$n_1$ and $n_2$ are the number of subjects with zero FRR and zero FAR for CASIA database. $n_3$ and $n_4$ are the number of subjects with zero FRR and zero FAR for ICE database. $n_5$ and $n_6$ are the number of subjects with zero FRR and zero FAR for MMU database. The listing of the number of subjects is given in Table 7.

Table 7. Number of subjects with zero FRR and FAR for CASIA, ICE and MMU databases

| Classifier | b | $n_1$ | $n_2$ | $n_3$ | $n_4$ | $n_5$ | $n_6$ |
|---|---|---|---|---|---|---|---|
| k-means (Euclidean) | 10 | 100 | 92 | 84 | 83 | 44 | 45 |
| | 8 | 85 | 92 | 78 | 73 | 39 | 45 |
| | 6 | 86 | 89 | 86 | 82 | 40 | 45 |
| | 3 | 89 | 87 | 86 | 82 | 43 | 43 |
| | 2 | 93 | 90 | 79 | 84 | 44 | 43 |
| k-means (Cityblock) | 10 | 98 | 90 | 79 | 80 | 41 | 44 |
|  |  | 8 | 87 | 91 | 78 | 80 | 38 | 43 |
|  |  | 6 | 88 | 92 | 77 | 79 | 40 | 42 |
|  |  | 3 | 88 | 86 | 81 | 80 | 42 | 41 |
|  |  | 2 | 91 | 97 | 82 | 82 | 40 | 41 |
| Fuzzy k-means |  | 10 | 89 | 88 | 84 | 84 | 43 | 44 |
|  |  | 8 | 88 | 87 | 77 | 78 | 38 | 41 |
|  |  | 6 | 88 | 89 | 80 | 81 | 39 | 42 |
|  |  | 3 | 89 | 87 | 86 | 81 | 43 | 41 |
|  |  | 2 | 92 | 91 | 83 | 82 | 41 | 42 |

The experiment was conducted on MCYT signature database of 75 subjects. Database consists of 15 genuine samples and 15 forged samples for each subject. The 10 genuine training samples per subject were considered after evaluating in rough set domain [22]. The distance measure used in k-means were Euclidean and city block distance. For fuzzy k-means, the distance measure considered was Euclidean distance. The experiment was conducted first with $Moment_A$ values for $b = 4,6,8$ and $d_1 = 128$. The difference in moment summation values for minimum variance blocks between genuine and imposter training samples were marginal. Slightly better performance was observed using $Moment_B$. The experiment was conducted for $b=4,6,8$ at $d_1 = 128$ and was extended to $b=8, 12, 16$ at $d_1=64$. Most promising results were obtained using MomentC and experiment was conducted for $b=4, 6, 8$ at $d_1=128$, $b=8, 12, 16$ at $d_1=64$ and $b=2, 3$ at $d_1=256$. The results are shown from Table 8-13. The trade-off between FRR and FAR were observed in the increase of $b$ value. The best performance of 8.13% of FRR and 9% of FAR at $d_1=128$ using $Moment_C$.

Table 8. $Moment_A$ FRR in % (at $d_1 =128$)

|  | b=4 | b=6 | b=8 |
|---|---|---|---|
| *k*-means(Euclidean) | 14 | 15.6 | 16 |
| *k*-means(Cityblock) | 13.6 | 14.8 | 15.7 |
| Fuzzy *k*-means | 12.9 | 14.2 | 15.6 |
| *k-nn* | 11.7 | 12.4 | 13.6 |
| Fuzzy *k-nn* | 12.2 | 13 | 13.4 |
| avg | 15.3 | 16.7 | 17 |
| avgmax | 13.2 | 14.6 | 15.6 |

Table 9. $Moment_A$ FAR in % (at $d_1 =128$)

|  | b=4 | b=6 | b=8 |
|---|---|---|---|
| *k*-means(Euclidean) | 14.1 | 13.8 | 13 |
| *k*-means(Cityblock) | 14.3 | 13.7 | 13 |
| Fuzzy *k*-means | 13.6 | 12.8 | 12.4 |
| *k-nn* | 13 | 12.6 | 12 |
| Fuzzy *k-nn* | 13.2 | 12.7 | 12.2 |
| avg | 14.1 | 13.6 | 13.1 |
| avgmax | 13.4 | 12.9 | 12.4 |

Table 10. Moment$_B$ FRR in % (columns 2-4 at $d_1 = 128$, columns 4-6 $d_1 = 64$)

|  | b=4 | b=6 | b=8 | b=8 | b=12 | b=16 |
|---|---|---|---|---|---|---|
| k-means(Euclidean) | 13.5 | 15.1 | 15.5 | 13.9 | 15.5 | 15.9 |
| k-means(Cityblock) | 13 | 14.2 | 15.1 | 13.3 | 14.5 | 15.4 |
| Fuzzy k-means | 12.3 | 13.6 | 15.0 | 12.8 | 13.6 | 15.0 |
| k-nn | 11.1 | 12.8 | 13.6 | 11.7 | 13.2 | 14.1 |
| Fuzzy k-nn | 12 | 12.6 | 13 | 12.4 | 13.0 | 13.6 |
| avg | 15 | 16.4 | 16.7 | 15.3 | 16.5 | 16.9 |
| avgmax | 13 | 14 | 15 | 13.5 | 14.6 | 15.6 |

Table 11. Moment$_B$ FAR in % (columns 2-4 at $d_1 = 128$, columns 4-6 $d_1 = 64$)

|  | b=4 | b=6 | b=8 | b=8 | b=12 | b=16 |
|---|---|---|---|---|---|---|
| k-means(Euclidean) | 13.7 | 13.4 | 12.7 | 14.2 | 13.8 | 13.2 |
| k-means(Cityblock) | 14 | 13.4 | 13 | 14 | 13.7 | 13.3 |
| Fuzzy k-means | 13.4 | 12.5 | 12.2 | 13.9 | 12.8 | 12.4 |
| k-nn | 13 | 12.4 | 12 | 13.3 | 12.5 | 12.2 |
| Fuzzy k-nn | 13.2 | 12.5 | 12.1 | 13.4 | 12.8 | 12.3 |
| avg | 14 | 13.4 | 13 | 14.4 | 13.7 | 13.5 |
| avgmax | 13.1 | 12.6 | 12.2 | 13.1 | 12.7 | 12.2 |

Table 12. Moment$_C$ FRR in % (columns 2-4 at $d_1 = 128$, columns 5-7 at $d_1 = 64$, columns 8-9 at $d_1 = 256$)

|  | b=4 | b=6 | b=8 | b=8 | b=12 | b=16 | b=2 | b=3 |
|---|---|---|---|---|---|---|---|---|
| k-means(Euclidean) | 10.7 | 12.2 | 12.5 | 10.9 | 12.5 | 13.0 | 13.2 | 14 |
| k-means(Cityblock) | 10.2 | 11.4 | 12.3 | 10.6 | 11.9 | 12.8 | 13 | 13.6 |
| Fuzzy k-means | 9.3 | 10.6 | 11.0 | 9.8 | 10.6 | 12.0 | 11.76 | 12.87 |
| k-nn | 8.13 | 9.82 | 10.61 | 8.71 | 10.21 | 11.12 | 11.5 | 12.5 |
| Fuzzy k-nn | 9 | 9.26 | 1 | 9.24 | 10.02 | 10.26 | 11.98 | 12.6 |
| avg | 12 | 13.4 | 13.7 | 12.3 | 13.5 | 13.9 | 13.65 | 14 |
| avgmax | 10 | 11 | 12 | 10.5 | 11.6 | 12.6 | 12.56 | 12.8 |

Table 13. Moment$_C$ FAR in % (columns 2-4 at $d_1 = 128$, columns 5-7 at $d_1 = 64$, columns 8-9 at $d_1 = 256$)

|  | b=4 | b=6 | b=8 | b=8 | b=12 | b=16 | b=2 | b=3 |
|---|---|---|---|---|---|---|---|---|
| k-means(Euclidean) | 10.36 | 10.2 | 9.7 | 11.2 | 10.6 | 10 | 12.7 | 11.10 |
| k-means(Cityblock) | 11.12 | 10.42 | 10 | 11.4 | 10.74 | 10.34 | 12.13 | 11.64 |
| Fuzzy k-means | 10.46 | 9.65 | 9.26 | 10.69 | 9.48 | 9.44 | 12.0 | 11.2 |
| *k-nn* | 10 | 9.4 | 9 | 10.3 | 10.5 | 10.23 | 11.0 | 10.5 |
| Fuzzy *k-nn* | 10.21 | 9.52 | 9.12 | 10.44 | 9.88 | 9.33 | 11.4 | 11.1 |
| avg | 11 | 10.42 | 10 | 11.44 | 10.78 | 10.35 | 12.23 | 11.9 |
| avgmax | 10.11 | 9.62 | 9.22 | 10.14 | 9.87 | 9.23 | 11.65 | 11 |

## 6. Conclusion

A new technique is presented which is simple and robust for authentication of off-line hand written signature using moment based descriptors. The technique uses Hu moments and is hence invariant to rotation, translation and scaling. The best performance was obtained for MMU iris database consisting of 45 subjects. The number of subjects with zero FRR was 44 and number of subjects with zero FAR was 45. False rejection rate of 8.13% and false acceptance rate of 10% was achieved on the MCYT database of 75 subjects, 30 samples each. Results show that performance is improved using k-nn classifier. Quadtree further depth decomposition will lead to more performance. The region quadtree can be easily extended to represent three dimensional data, leading to octree. The third dimension can be a static feature of the signature and can be classified using continuous dynamic programming [20].

## Acknowledgements

The authors would like to thank J.Ortega-Garcia for the provision of MCYT Signature database from Biometric Recognition Group, B-203, Universidad Autonoma de Madrid SPAIN [21]. The authors thank Patrick Flynn from University of NotreDame, USA, for ICE database, Chinese Academy of Sciences Institute of Automation for CASIA database and Multimedia University for MMU database.

## References

[1] Gady Agam, Suneel Suresh, "Warping-based offline signature recognition", in Proc. IEEE Transactions on Information forensics and security, Vol. 2, No. 3, pp. 430-437, 2007.

[2] George D. da C. Cavalcanti, Rodrigo C. Doria and Edson C. de B.C. Filho, "Feature selection for off-line recognition of different size signatures", in Proc. 12th IEEE workshop on Neural Network for Signal Processing, Vol. 1, pp. 355-364, 2002.

[3] G.K.Gupta, R.C.Joyce, "Using position extrema points to capture shape in on-line hand written signature verification", Pattern Recognition, Vol. 40, pp. 2811-2817, 2007.

[4] Thierry Denceux, "A k-Nearest Neighbor Classification Rule Based on Dempster-Shafer Theory", IEEE Transactions on Systems, Man and Cybernetics, Vol. 25, No. 5, pp. 804-813, 1995.

[5] V. Dasarathy, "Minimal Consistent Set (MCS) Identification for Optimal Nearest Neighbor Decision Systems Design", IEEE Tranactions on Systems, Man and Cybernetics, Vol. 24, No. 1, pp. 511-517, 1994.

[6] S. Ferilli M. Biba T.M.A. Basile N. Di Mauro, "k-Nearest Neighbor Classification on First-Order Logic Descriptions", IEEE International Conference on Data Mining Workshops, pp. 202-210, 2008.

[7] Anil K. Ghosh, Probal Chaudhuri, and C.A. Murthy, "On Visualization and aggregation of Nearest Neighbor Classifiers", IEEE Transactions on Pattern Analysis and Machine Intelligence, Vol. 27, No. 10, pp. 1592-1602, 2005.

[8] Yongguang Bao, Xiaoyong Du, Naohiro Ishii, "Improving Performance of the K-Nearest Neighbor Classifier by Tolerant Rough Sets", in Proc. Third International Symposium on Cooperative Database Systems for Advanced Applications, pp. 167-171,2001.

[9] Lei Wang, Latifur Khan and Bhavani Thuraisingham,"An Effective Evidence Theory based K-nearest Neighbor (KNN) classification", IEEE/WIC/ACM International Conference on Web Intelligence and Intelligent Agent Technology, pp. 797-801, 2008.

[10] Strzecha K., "Image segmentation algorithm based on statistical pattern recognition methods", in Proc. 6th Int. Conf. on The Experience of Designing and Application of CAD Systems in MicroElectronics, pp. 200-201, 2001.

[11] K.R. Radhika, M.K. Venkatesha, G.N. Sekhar, "On-line Signature Authentication using Zernike moments", Proc. IEEE Int Conf on Biometrics: Theory, Applications and Systems, 2009, pp. 1-4.

[12] M. Hu, "Visual pattern recognition by moment invariants", IRE Transactions on Information Theory, Vol. 8, pp. 179-187, 1962.

[13] Foster J., Nixon M. S. and Prugel-Bennett A., "Gait Recognition by Moment Based Descriptors", 4th Int. Conf. Recent Advances in Soft Computing, pp. 78-84, 2002.

[14] Vu Nguyen, Michael Blumenstein, Vallipuram Muthukkumarasamy, Graham Leedham, "Off-line signature verification using enhanced modified direction features in conjunction with neural classifiers and support vector machines", in Proc. Int Conf of Pattern Recognition, Vol. 4, pp. 509-512, 2006.

[15] Manuel R. Freire, Julian Fierrez, Marcos Martinez-Diaz, Javier Ortega-Garcia, "On the applicability of off-line signatures to the fuzzy vault construction", in Proc. Int Conf on Document Analysis and Recognition, Vol. 1, pp. 1173-1177, 2007.

[16] Edson J.R.Justino, Flavio Bortolozzi, Robert Sabourin, "A comparison of SVM and HMM classifiers in the off-line signature verification", Pattern Recognition Letters, Vol. 26, pp. 1377-1385, 2004.

[17] Madasu Hanmandlu, Mohd. Hafizuddin Mohd Yusof, Vamsi Krishna madasu, "Off-line Signature verification and forgery detection using fuzzy modeling", Pattern Recognition, Vol. 38, pp. 341-356, 2004.

[18] Zeng Yong, Wang Bing, Zhao Liang, Yang Yupu,"The Extended Nearest Neighbor Classification", in Proc. 27th Chinese Control Conference, pp. 559-563, 2008.

[19] K.R.Radhika, S.V.Sheela, M.K.Venkatesha and G.N.Sekhar, "Multi-modal Authentication using Continuous Dynamic Programming", International Conference on Biometric ID Management and Multimodal Communication, Springer LNCS Vol. 5707, 2009, pp. 228-235.

[20] J.Ortega-Garcia, Fierrez-Aguilar et al, "MCYT Baseline Corpus: A Bimodal Biometric Database", IEEE Proc. Vision, Image and Signal Processing, vol. 150, pp. 395-401, 2003.

[21] Rose 2 Rough Sets Data Explorer Version 2.2, http://wwwidss.cs.put.poznan.pl/rose.

[22] Rafeal C. Gonzalez, Richard E. Woods, "Digital Image Processing", Second Edition, Pearson Education, 2005.

[23] T. H.Reiss, "The Revised Fundamental Theorem of Moment Invariants", IEEE Trans. on Pattern Analysis and Machine Intelligence, Vol. 13, No. 8, 1991.

[24] J. Daugman, High Confidence Visual Recognition by a Test of Statistical Independence, IEEE Trans. Pattern Analysis and Machine Intelligence, Vol. 15, No.11, pp.1148-1161,1993.

[25] Karen Hollingsworth, Sarah Baker, Sarah Ring, Kevin W. Bowyer and Patrick J. Flynn, "Recent Research Results In Iris Biometrics", SPIE 7306B: Biometric Technology forHuman Identification VI, 2009.

[26] R. Wildes, J. Asmuth, G. Green, S. Hsu, R. Kolczynski, J. Matey, and S. McBride, A machine-vision system for iris recognition, Mach. Vis. Applic., vol. 9, pp. 1-8, 1996.

[27] Li Ma, Tieniu Tan, Yunhong Wang, Dexin Zhang, Personal Identification based on Iris Texture Analysis, IEEE Transactions on Pattern Analysis and Machine Intelligence, Vol.25, No.12, December 2003.

[28] Zhenan Sun, Yunhong Wang, Tieniu Tan and Jiali Cui, "Improving Iris Recognition Accuracy Via Cascaded Classifiers", Biometric Authentication. Vol. LNCS 3072, pp. 1-11, 2004.

[29] H. Proença and L.A. Alexandre, "Iris Segmentation Methodology for Noncooperative Iris Recognition," IEE Proc. Vision, Image, and Signal Processing, vol. 153, no. 2, pp. 199-205, Apr. 2006.

[30] CASIA-IrisV3, http://www.cbsr.ia.ac.cn/IrisDatabase.htm

[31] Xiaomei Liu, Bowyer K.W., Flynn, P.J., "Experimental Evaluation of Iris Recognition", IEEE Computer Society Conference on Computer Vision and Pattern Recognition Workshops, pp. 158-165, 2005.

[32] MMU database http://www.pesona.mmu.edu.my

[33] Nicolaie Popescu-Bodorin, "A fuzzy view on k-menas based signal quantization with application in iris segmentation", 17[th] Telecommuniations Forum, pp 1-4, 2009.

[34] W.S.Chen. S.W.Shih, K.T.Shen, L.Hsieh and K.H.Chih, "Automatic iris recognition system based on wavelet transform and energy transform", 15[th] IPPR Conference on Computer Vision, Graphics and image Processing, pp. 25-27, 2002.

**Authors**

S.V.Sheela is working as a faculty in the department of Information Science and Engineering since 1999. She received her Master of Engineering degree in Computer Science and Engineering in 2004. She is working as Senior Lecturer, in the department of Information Science and Engineering, B.M.S College of Engineering, Bangalore, Karnataka, India. She is currently working on iris recognition techniques. She is a member of IEEE.

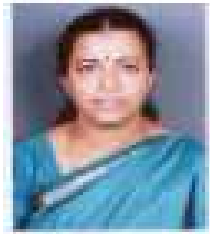

Radhika K.R is working as a faculty, since 1995 in the department of Information Science and Engineering. She received her Master of Engineering degree in Computer Science in 2000. Presently, she is working as Assistant Professor, in the department of Information Science and Engineering, B.M.S College of Engineering, Bangalore, Karnataka, India. She is currently, working on Online and Offline Biometric Behavioral Pattern Authentication System (BBPAS) using handwritten signature as a pattern. She is a member of IEEE and the IEEE Computer Society.

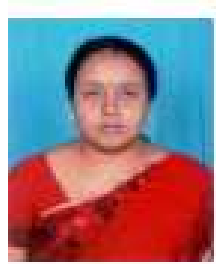